\documentclass{article}

 \usepackage[final]{neurips_mlsb_2025}

\usepackage[utf8]{inputenc} 
\usepackage[T1]{fontenc}    
\usepackage{hyperref}       
\usepackage{url}            
\usepackage{booktabs}       
\usepackage{amsfonts}       
\usepackage{nicefrac}       
\usepackage{microtype}      
\usepackage{xcolor}         

\usepackage{amsmath,amsfonts,bm}

\def\eqref#1{equation~\ref{#1}}

\def\1{\bm{1}}

\def\vw{{\bm{w}}}
\def\vx{{\bm{x}}}

\DeclareMathAlphabet{\mathsfit}{\encodingdefault}{\sfdefault}{m}{sl}
\SetMathAlphabet{\mathsfit}{bold}{\encodingdefault}{\sfdefault}{bx}{n}

\usepackage{graphicx}
\usepackage{wrapfig}
\usepackage{xcolor}
\usepackage{enumitem}

\newcommand{\tbf}[1]{\textbf{#1}}

\title{
Conformational Rank Conditioned Committees for Machine Learning-Assisted Directed Evolution
}

\author{
Mia Adler\thanks{Equal contribution; sorted alphabetically.} \\ Pomona College\And
Carrie Liang$^{*}$\\ The University of Texas at Austin \And
Brian Peng$^{*}$ \\ University of Cambridge \And
Oleg Presnyakov$^{*}$ \\University of California, Irvine \And
Justin Baker\thanks{Correspond to \texttt{justin@math.ucla.edu}}\\
University of California, Los Angeles\\
 \And 
Jannelle Lauffer,
Himani Sharma,
Barry Merriman\\
Avery Bio Inc.
}

\begin{document}

\maketitle

\begin{abstract}
Machine Learning-assisted directed evolution (MLDE) is a powerful tool for efficiently navigating antibody fitness landscapes.
Many structure-aware MLDE pipelines rely on a single conformation or a single committee across all conformations, limiting their ability to separate conformational uncertainty from epistemic uncertainty.
Here, we introduce a rank‑conditioned committee (RCC) framework that leverages ranked conformations to assign a deep neural network committee per rank.
This design enables a principled separation between epistemic uncertainty and conformational uncertainty.
We validate our RCC-MLDE approach on SARS-CoV-2 antibody docking, demonstrating significant improvements over baseline strategies.
Our results offer a scalable route for therapeutic antibody discovery while directly addressing the challenge of modeling conformational uncertainty.
\end{abstract}

\maketitle


\vspace{-0.3cm}
\section{Introduction}\label{sec:introduction}
\vspace{-0.3cm}
Antibody therapeutics have recently become the dominant type of drug, overtaking small molecules.
As a result, it is of great interest to develop methods to design antibodies to bind a given target antigen.
Here we develop a computational pipeline for antibody design optimization that makes use of state-of-the art existing techniques while explicitly addressing conformational uncertainty.
The algorithms are iterative by design, so that they can in the future be naturally used in a {\em lab-in-the-loop} format, to engage directly with experimental exploration of designs.

Machine learning-assisted directed evolution (MLDE) has significantly enhanced therapeutic antibody design by accelerating the exploration of vast sequence spaces
and prioritizing high-affinity variants~\cite{qiu2021cluster,yang2025active}.
In parallel, modern folding algorithms and ensemble techniques now rapidly generate accurate protein conformations for use in {\em in silico} antibody optimization~\cite{abanades2023immunebuilder,jingAlphaFoldMeetsFlow2024}.
Typical {\em in silico} antibody binding pipelines perform an initial stage of rigid‑body docking followed by a semi‑flexible pose refinement at the side chain and backbone interface.
However, large loop rearrangements are still hard to realize unless present in the starting ensemble~\cite{honorato2024haddock2}.
Flexible docking alternatives can remodel paratopes, but their performance remains sensitive to input ensembles and computationally demanding.
Thus, rich conformational ensembles are critical for computationally efficient antibody optimization~\cite{bhattTeachingOldDocks2024}.
This is particularly important in the case of SARS-CoV-2, where state switching in the receptor binding domain (RBD)~\cite{yajimaStructuralBasisReceptorbinding2024} and variability in
the heavy-chain complementarity-determining region (CDR-H3)~\cite{barnesStructuresHumanAntibodies2020, asarnowStructuralInsightSARSCoV22021} complicate binding affinity prediction~\cite{abanades2023immunebuilder}.
Although ensemble techniques may broaden structural coverage, the critical difficulty in combining these approaches with MLDE lies in disentangling uncertainty arising from binding pose selection from uncertainty in the accuracy of the model~\cite{kendallWhatUncertaintiesWe2017}.
Addressing this separation is essential for antibody design, especially against novel pathogens that stretch our current understanding and may drive future outbreaks and pandemics.

In protein design, structure-aware MLDE often couples surrogate models with Bayesian active learning to propose new sequences using acquisition functions such as expected improvement or upper confidence bounds that depend on both the surrogate mean and predictive uncertainty~\cite{yang2025active}.
A key challenge is that predictive variance can originate from two conceptually distinct sources: epistemic uncertainty due to model limitations and aleatoric uncertainty arising from conformational heterogeneity of antibody–antigen complexes~\cite{kendallWhatUncertaintiesWe2017}.
Standard ensemble approaches such as deep ensembles generally conflate these effects.
Bayesian Committee Machines provide principled ways to combine estimators~\cite{trespBayesianCommitteeMachine2000}, and recent work has explored ensemble learning over conformational docking ensembles~\cite{mohammadiEnsembleLearningEnsemble2022}.
Building on these ideas, our contribution is to introduce a rank-conditioned committee that stratifies ensembles by conformation rank, so that within-rank dispersion captures epistemic variance while between-rank dispersion captures aleatoric variance, thereby enabling acquisition functions that more reliably balance exploration and exploitation in MLDE.
\begin{figure}[!ht]
\vspace{-.3cm}
\centering
\begin{tabular}{ccc}
        \hspace{-.25cm}
        \includegraphics[width=0.5\textwidth, clip, trim = 0cm .2cm 0cm 0cm]{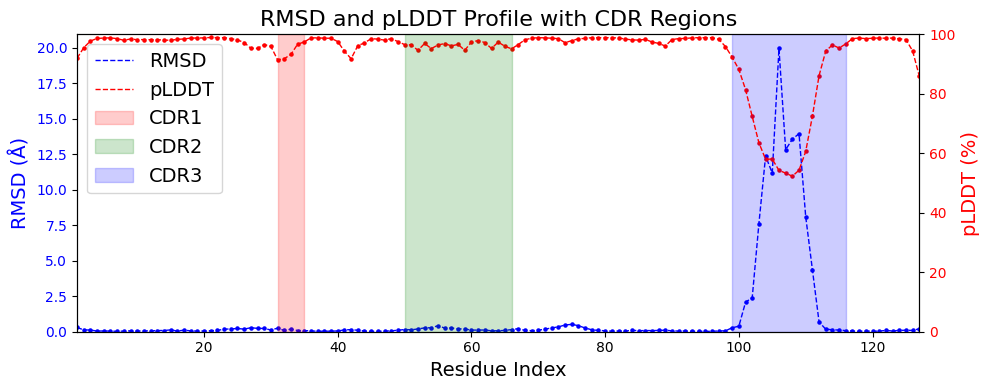}
    \includegraphics[width=0.21\textwidth, clip, trim = 2cm 0cm 1.5cm 0cm]{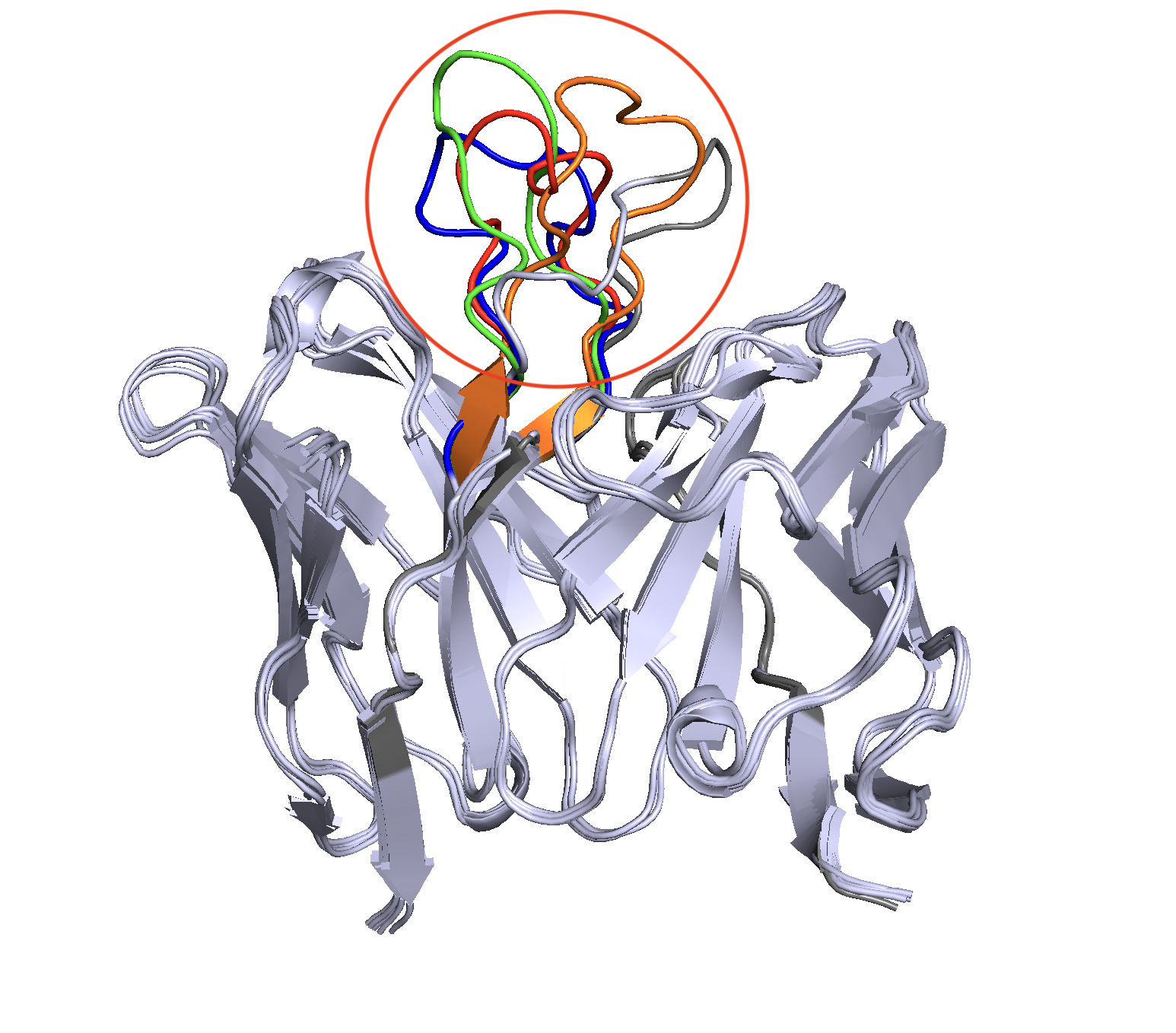}
    & \hspace{-.8cm}
    \includegraphics[width=0.29\textwidth, clip, trim = 0cm -2.5cm 0cm 0cm]{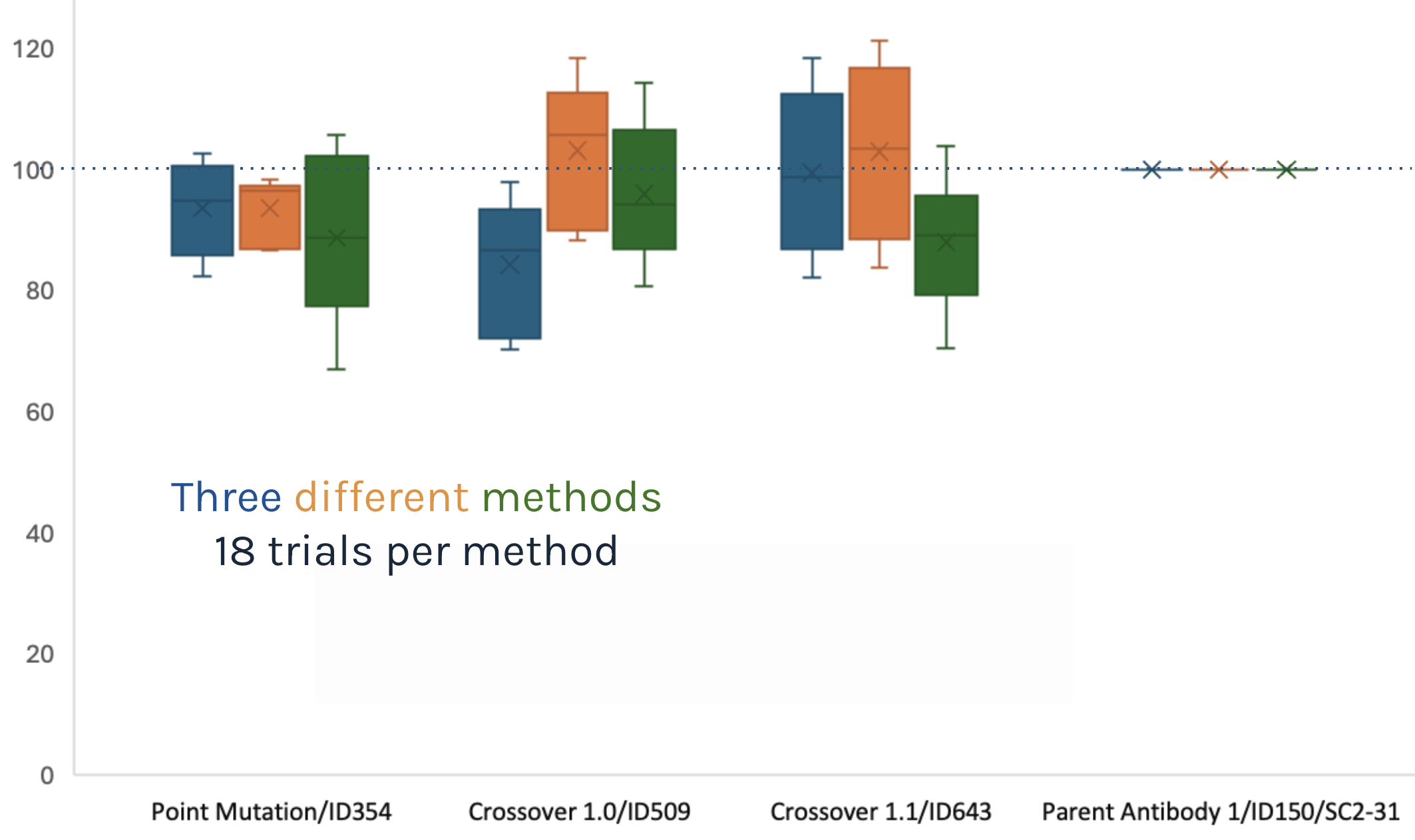}
\end{tabular}
 
\caption{
The \tbf{left} panel depicts the residue-wise confidence scores in AlphaFold2 and ImmuneBuilder.
The \tbf{middle} panel shows {five} {\em aligned} conformal antibodies with predicted by AlphaFold2 and ImmuneBuilder with colored CDR-H3 regions circled in red.
The \tbf{right} panel presents wet-lab results of the best performing antibodies in three consecutive rounds of DE, normalized to the parent.
}
\label{fig:ConformalFlex}
\vspace{-.3cm}
\end{figure}


Figure~\ref{fig:ConformalFlex} illustrates the practical challenges of modeling antibodies under conformational uncertainty.
In the left panel, the residue-wise confidence scores for AlphaFold2~\cite{AlphaFold2021} and ImmuneBuilder~\cite{abanades2023immunebuilder} are reported, with CDR-H3 showing significantly lower confidence relative to the framework regions.
The middle panel shows the conformations predicted by AlphaFold2 and Immunebuilder.
Each of these sequences are aligned and their CDR-H3 regions are colored and circled.
There is a nearly exact match outside of the CDR-H3 region, but significant variance among the CDR-H3 loops themselves.

The right panel of Figure~\ref{fig:ConformalFlex} connects these modeling challenges to experimental outcomes for the best performing antibodies after five rounds of biologically informed directed evolution. The experimental method is described in detail in \cite{huntRapidCellfreeExpression2023}. Briefly, DNA genes for variable heavy and light-chain sequences prodcued by the AI antibody pipeline were designed in SnapGene~\cite{snapgene}, synthesized as gBlocks (IDT),  and amplified and purified for direct use in cell-free protein synthesis (CFPS).
CFPS was carried out with the NEB PURExpress kit.
Binding of expressed single domain antibody (sdFab) constructs to the target antigen (SARS-CoV-2 Receptor Binding Domain (RBD)) was evaluated using a solution-phase flourescent reporter assay for binding ( AlphaLISA assay~\cite{bielefeld2009alphalisa}) : CFPS products were incubated with the RBD antigen, followed by AlphaLISA donor/acceptor bead addition, and reporter luminescence was measured on a plate reader.
Signals were background-subtracted, averaged across replicates, and normalized to a literature parent antibody (ID150/SC2-31).
Positive controls yielded strong Alpha signals while negative controls remained at baseline, confirming assay specificity.
The figure shows the results for three different antibody designs, as normalized to the parent antibody results (each design is tested in independent triplicates (shown as three colored bars) with 6 technical replicates each (shown as error bars)), and these show a trend that the improved designs do indeed tend to produce better measured binding (with the caveats of experimental variability, and that the AlphaLISA assay is not a direct measurement of equilibrium binding affinity).

\vspace{-0.3cm}
\section{Machine Learning Directed Evolution (MLDE)}\label{sec:bkgnd}
\vspace{-0.3cm}

One key element of antibody design is to optimize the binding affinity for the target.
In this context, the antibody design problem can be formulated as the constrained optimization problem
\[
  \arg\min_{s\in S} h(s),
\]
where \( S \) is the set of all developable antibody sequences, and \( h(s) \) is the performance metric, which in this case is the HADDOCK3~\cite{honorato2024haddock2} docking score between a given antibody sequence \( s \) and the fixed SARS-CoV-2 antigen.
The CDR-H3 region of a given sequence $s$ includes at least 30 mutable residues, each of which can take one of 20 amino acids, yielding a sequence space of size $|S| \geq 20^{30}$.
Exhaustive search over such a space is infeasible.
Moreover, evaluating the objective function $h(s)$ is computationally expensive, limiting the number of sequences and poses that can be directly assessed.
 
Rather than attempting exhaustive search, it is natural to solve this optimization problem through via iterative evolution of the design,  using {\em in silico} or experimental evaluations of intermediate candidates. 
Such strategies leverage the demonstrated power of directed evolution methodologies.
In particular,
MLDE addresses these challenges by replacing exhaustive search with iterative rounds of surrogate modeling and sequence selection.
The success of MLDE depends critically on strategies that (i) reduce the effective search space by incorporating biologically informed constraints, and (ii) improve sampling efficiency through Bayesian active learning.
This section reviews these complementary approaches and their role in enabling tractable antibody optimization.


\subsection{Biologically Informed Directed Evolution}
\label{subsec:directed-evolution}
In classical directed evolution (DE) pipelines, new antibody variants are generated by introducing mutations into existing sequences.
These mutations can take the form of residue substitutions or cross-over mutations, and in purely random DE they are typically applied uniformly across the sequence.
For every new variant, mutation type with equal probability,
\[
\mathbb{P}[\text{substitution}] = \mathbb{P}[\text{crossover}] = 0.5,
\]
to ensure balanced exploration.

To better reflect biological constraints, substitution mutations can instead be biased toward substitutions that are both more likely to occur in nature and more likely to affect functionally critical regions.
We obtain such biases by performing sequence alignment across related variants, which highlights substitutions enriched in homologous antibodies and thereby informative for antigen binding.
This procedure defines a biologically informed substitution matrix, which specifies position- and residue-dependent mutation weights and is described in detail in Appendix~\ref{sec:Bioinformatics1}.
Formally the matrix 
\[
P \in [0,1]^{|\mathcal{A}|\times|\mathcal{A}|}, \quad \sum_{b\in\mathcal{A}} P_{ab} = 1,
\]
encodes the probability \(P_{ab}\) of substituting amino acid \(a\) with \(b\).

In the case of a crossover mutation, two parent sequences \(s_p\) and \(s_q\) are drawn from the current population according to a weighted probability
\[
\mathbb{P}[s] \propto {g(s) ^ 2}, \quad g(s)=-h(s),
\]
favoring high-performing binders, and a uniform cut position \(k \sim \mathrm{Unif}\{1,\dots,L-1\}\) is chosen to form the recombinant
\[
c = (s_p[1{:}k],\, s_q[k{+}1{:}L]),
\]
reflecting the drastic nature of this operation. For a substitution mutation, a single parent \(s_p\) is selected from the same weighted distribution, a CDR region \(R\) is chosen, and a residue \(a=(s_p)_i\) at position \(i \in \mathcal{I}_R\) is replaced by a new residue \(a'\) drawn from the categorical distribution defined by \(P_{a,\cdot}\).
 In both cases, the selection pressure through \(\Pr_t(s)\) biases the generation toward variants derived from sequences with superior docking scores, ensuring that exploration is guided by prior performance.

\subsection{Bayesian Active Learning} 
\label{sec:bo}
In MLDE, only a limited number of sequences can be experimentally or computationally evaluated, making it essential to identify candidates that provide the most informative feedback for model improvement.
Bayesian active learning addresses this challenge by employing probabilistic surrogate models whose predictive distributions quantify both expected performance and uncertainty.
The process proceeds in two steps: first, antibody sequences are embedded into a continuous feature space, and second, active learning strategies are applied in this space to select sequences that balance exploration and exploitation.

\paragraph{Protein Embedding}
Antibody sequence embeddings leverage the numerical and analytic tractability of real vector spaces to represent antibodies.
We may denote this embedding as a mapping
$\iota:S\to\mathbb R^d.$
Among many types of embeddings, it is critical to ensure that the embeddings capture the structural and functional properties of the protein molecules they represent.
For example, traditional one-hot encoding faces the challenge of high and variable dimensions, as an alphabet of size 20 and a sequence length greater than 100 yield an embedding of variable dimension over 2000.

To obtain a fixed-and-low-dimensional embedding, we apply Antibody Mutagenesis-Augmented Processing (AbMAP), a framework introduced by Singh et al. \cite{singh2025learning}.
 AbMAP begins with representations from foundational protein language models (PLMs), which mostly encode information across conserved regions of protein sequences.
 To extract embedded information from antibody hypervariable CDR regions, AbMAP also generates embeddings for randomized mutants of the input sequence for comparison.
 By subtracting the average mutant embedding from the original, the framework isolates CDR-specific contributions, a curated embedding that is passed through a downstream transformer-based neural network to yield fixed-dimensional representations tailored to the structural and functional landscape of antibody hypervariable regions.

\paragraph{Active Learning}
In MLDE, the role of active learning is to decide which unevaluated sequences should be tested next, so that each round of data collection maximally improves the surrogate model.
At iteration 
$n$, we maintain a dataset
\[
D_n = \{(\vx, h(\vx)) \mid \vx = \iota(s),\, s \in S_n\}, \quad S_n \subseteq S,
\]
where $h(\vx)$ denotes the measured property (e.g., binding score) and $\iota(s)$ embeds sequence $s$ into the feature space.
To estimate predictive uncertainty, we adopt an ensemble active learning strategy, training a committee of $M$ independent surrogate models
\(
\{\hat{h}_1, \ldots, \hat{h}_M\}
\)
on bootstrap resamples of $D_n$.
For any candidate sequence $s \in S \setminus S_n$ with embedding $\vx=\iota(s)$, we summarize the committee with mean and variance
\[
\hat{\mu}(\vx) = \frac{1}{M}\sum_{m=1}^{M} \hat{h}_m(\vx), \qquad
\hat{\sigma}(\vx) = \sqrt{\frac{1}{M-1} \sum_{m=1}^{M} \big(\hat{h}_m(\vx) - \hat{\mu}(\vx)\big)^2}.
\]

An acquisition function $\alpha(\vx)$ then maps these predictions into a scalar score that balances \emph{exploitation} (selecting sequences with high predicted performance $\hat{\mu}(\vx)$) against \emph{exploration} (selecting sequences with high uncertainty $\hat{\sigma}(\vx)$). A standard choice is the \emph{upper confidence bound (UCB)}
\[
\alpha(\vx) = \hat{\mu}(\vx) + \kappa \, \hat{\sigma}(\vx),
\]
where $\kappa > 0$ controls the trade--off. Larger values of $\kappa$ encourage exploratory sampling of uncertain candidates, while smaller values favor exploitation of high--scoring predictions. This mechanism prioritizes sequences that are both promising and informative, ensuring efficient navigation of the antibody sequence space.

\vspace{-0.3cm}
\section{Modeling Uncertainty via Rank-Conditioned Committees}\label{sec:methods}
\vspace{-0.3cm}

The key question of the active learning approach is how to handle conformational ensembles.
If one trains a single model over an ensemble of conformations and their docking scores, then the uncertainty is tied to the docking uncertainty and not to model uncertainty.
If one trains a committee over multiple conformations, then the uncertainty is conflated.
This leads us to design rank-conditioned committees, a mixture of committees that disentangles the uncertainty between the committee predictions and the uncertainty in the conformational pose.

\subsection{Ensemble Based Acquisition Maximization}

Our approach combines both active learning and directed evolution.
At iteration $n$,  let $r\in\{1,\ldots,R\}$ index folding-derived ranks, e.g., top-$R$ ImmuneBuilder conformations.
For each rank $r$, we train a committee of $M$ surrogates on the rank-specific dataset $D^{(r)}_n=\{(\vx_i,y_{i,r})\}$ where $y_{i,r}$ is the docking score of the sequence $s_i$ with fold rank $r$.
Given $\vx$ the committee produces a rank-conditional mean and epistemic standard deviation
\[
\hat\mu_r(\vx)\;=\;\frac{1}{M}\sum_{m=1}^{M}\hat h_{r,m}(\vx),\qquad
\hat\sigma_{\text{epi},r}(\vx)\;=\;\sqrt{\frac{1}{M-1}\sum_{m=1}^M\!\big(\hat h_{r,m}(\vx)-\hat\mu_r(\vx)\big)^2}.
\]

To aggregate over conformations while separating uncertainty sources, we use weights $\vw=(w_1,\dots,w_R)$ with $\sum_r w_r=1$, e.g., uniform over the top $R$ ranks or a calibrated rank prior. The RCC approach allows us to define the sequence--level statistics
\[
\bar\mu(\vx)\;=\;\sum_{r=1}^R w_r\,\hat\mu_r(\vx),\,\,\,\,
\hat\sigma_{\text{epi}}^2(\vx)\;=\;\sum_{r=1}^R w_r\,\hat\sigma_{\text{epi},r}^2(\vx),\,\,\,\,
\hat\sigma_{\text{conf}}^2(\vx)\;=\;\sum_{r=1}^R w_r\big(\hat\mu_r(\vx)-\bar\mu(\vx)\big)^2,
\]
so that the total predictive variance decomposes into epistemic and conformational components
\[
\hat\sigma_{\text{tot}}^2(\vx)\;=\;\hat\sigma_{\text{epi}}^2(\vx)+\hat\sigma_{\text{conf}}^2(\vx).
\]

A candidate library $C_n$ is a set of unscored sequences generated via one round of directed evolution, using stochastic mutation matrices with equal off-diagonal transition probabilities to promote diversity.
For each candidate sequence $s\in C_n$ with embedding $\vx=\iota(s)$, we predict performance using the rank-conditioned committees.
The candidates are then scored with a rank--conditioned acquisition function; a simple two--term variant is
\[
\alpha_{\text{RCC}}(\vx)\;=\;\bar\mu(\vx)\;+\;\kappa_{\text{epi}}\hat\sigma_{\text{epi}}(\vx)\;-\;\kappa_{\text{conf}}\hat\sigma_{\text{conf}}(\vx),
\]
where $\kappa_{\text{epi}}>0$ promotes exploration driven by model uncertainty, while $\kappa_{\text{conf}}\ge 0$ optionally down--weights candidates whose uncertainty is predominantly conformational.
Special cases recover standard practice, e.g., $\kappa_{\text{conf}}{=}0$ yields UCB on the rank--averaged predictor.

We compute $\alpha_{\text{RCC}}(\vx)$ for all $\vx\in C_n$ and select the top $B$ sequences,
\[
S_n^\star \;=\; \operatorname{Top}_B\{\alpha_{\text{RCC}}(\vx)\mid \vx\in C_n\}.
\]

For each $x\in S_n^\star$, we perform docking to obtain rank--specific labels $\{y_{r}(x)\}_{r=1}^R$ and update all rank datasets,
\[
D_{n+1}^{(r)} \;=\; D_n^{(r)} \,\cup\, \{(\,x,\, y_r(x)\,)\}\qquad \text{for } r=1,\dots,R.
\]
The process is iterative, where each rank-conditioned committees can be retrained and new candidate libraries can be generated.

\subsection{Methods}

We use the SARS‑CoV‑2 antibody database~\cite{hunt2023rapid}, and build a fully autonomous, end‑to‑end \textit{in silico} pipeline by integrating open‑source tools (Riot‑NA~\cite{riot}, ImmuneBuilder~\cite{abanades2023immunebuilder}, PBDFixer~\cite{pdbfixer}, and HADDOCK3~\cite{honorato2024haddock2}).
ImmuneBuilder\cite{abanades2023immunebuilder}, which is specialized for antibodies, returns four predicted conformations per sequence in PDB format.
Per‑residue variability across these predicted conformations is summarized via RMSD to capture local structural uncertainty. 
Docking is performed using HADDOCK3 \cite{honorato2024haddock2}, a data-driven platform for modeling biomolecular complexes.
HADDOCK3 integrates experimental data and physicochemical parameters to generate structural models of protein-to-protein interactions.
For each antibody-antigen pair, the software produces multiple docked complexes in PDB format, along with detailed scoring information that includes van der Waals, electrostatics, desolvation energy, and an overall HADDOCK binding score.
The final HADDOCK binding score is a weighted sum of energy terms, designed to approximate the quality and stability of the predicted protein–protein complex.
A more negative score indicates stronger binding for a given complex, which is why we employ a minimization function. HADDOCK3 groups resulting models into clusters based on structural similarity, and only top-scoring clusters are retained for downstream analysis. This procedure enables us to prioritize variants with strong predicted binding ability and plausible interaction interfaces.

\paragraph{Our RCC-MLDE Approach.}
The RCC-MLDE framework runs in sequential rounds. 
At iteration $n$, we maintain a labeled set 
$D_n = \{(x_i, \{y_{i,r}\}_{r=1}^R)\}$ 
of all sequences $s_i$ that have been evaluated so far, where 
$x_i = \iota(s_i)$ is the AbMAP embedding and $y_{i,r}$ is the HADDOCK3 docking score for each ImmuneBuilder conformational rank $r$. 
From the current population, we generate a candidate library $C_n$ via bioinformatics-guided single- and double-site substitutions and parental crossovers, then filter these sequences for basic biological feasibility (Riot-NA) and structural plausibility. 
For each rank $r$, we construct a rank-specific dataset 
$D_n^{(r)} = \{(x_i, y_{i,r})\}$ 
and train an ensemble (``committee'') of surrogate models, either fully connected DNNs or XGBoost regressors, on $D_n^{(r)}$. 
Given a candidate embedding $x \in C_n$, each committee produces a rank-conditional mean and epistemic variance, which we aggregate across ranks either with a standard ensemble acquisition (baseline MLDE) or with the rank-conditioned committee acquisition defined in Section~3.1 that separately tracks epistemic and conformational uncertainty. 
We then score all $x \in C_n$, select the top $B$ sequences under the chosen acquisition, run ImmuneBuilder and HADDOCK3 to obtain new labels $\{y_{r}(x)\}_{r=1}^R$, and form $D_{n+1}$ by appending these measurements. 
There is no static train/test split; at each iteration models are trained on $D_n$ with bootstrapping and evaluated only on unseen candidates in $C_n$, and newly evaluated sequences are used for training only in subsequent rounds. 
We depict this pipeline in Appendix~\ref{app:exp}, and our experimental results summarize the distribution of docking scores of the new antibodies.

\vspace{-0.3cm}
\section{Experimental Results}
\vspace{-0.3cm}

In our experiments, we compared three main approaches: (i) bioinformatics-based directed evolution, (ii) machine learning-assisted directed evolution (MLDE) with an XGBoost ensemble, and (iii) MLDE with a deep neural network (DNN) ensemble.
The primary objective was to evaluate whether our methods can consistently improve the fitness landscape by identifying higher-scoring antibody variants.
To this end, we tracked the mean score of the population across multiple runs.
An upward shift in the mean score indicates that the approach is effective in discovering beneficial variants.
Moreover, the magnitude of this shift reflects the efficiency of the method: the larger and faster the improvement, the more effective the strategy is at navigating the search space.
Additionally, variance plays a crucial role in evaluating the algorithm, as it reflects the diversity and explorability of the search.
Algorithms that achieve a large mean shift while maintaining moderate variance are prioritized, as they demonstrate both effectiveness and the ability to explore diverse regions of the solution space.

\begin{figure}[!ht]
\centering
    \begin{tabular}{cc}
\includegraphics[width=0.45\textwidth]{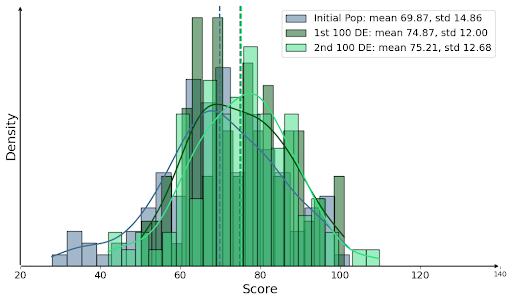}
&
\includegraphics[width=0.45\textwidth]{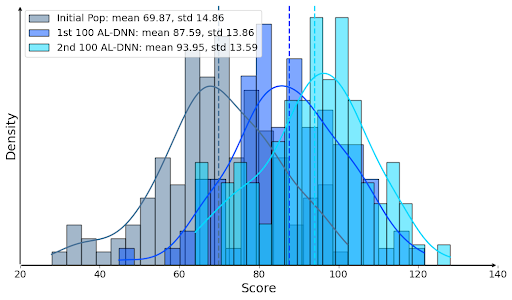}
\\
(a) Directed Evolution & (b) Baseline MLDE with DNN \\
\includegraphics[width=0.45\textwidth]{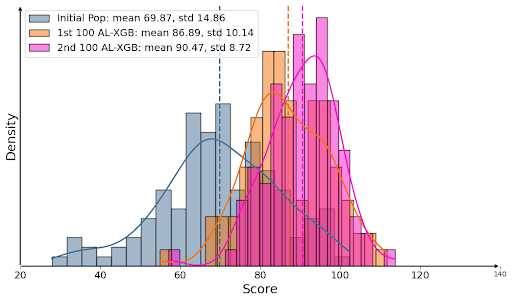}
&
\includegraphics[width=0.45\textwidth]{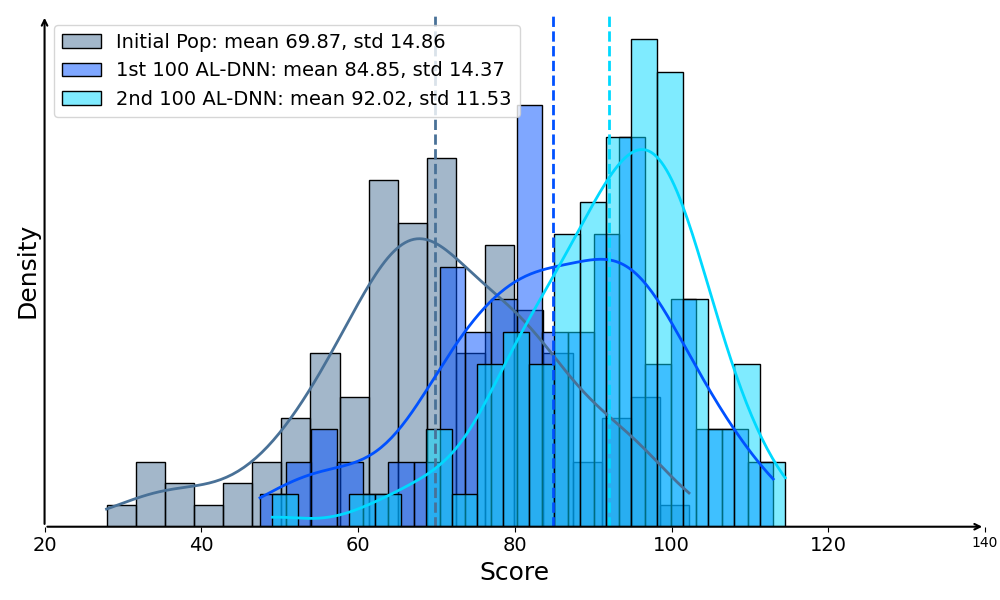}\\
(c) RCC MLDE with XGBoost & (b) RCC MLDE with DNN \\
    \end{tabular}
\caption[Comparison of results]{Comparison of results for (a) bioinformatics-based DE (b) MLDE with a DNN ensemble (c) RCC-MLDE with XGBoost and (d) RCC-MLDE with a DNN ensemle. Each histogram shows the evaluation of the initial dataset (grey) alongside 200 generated antibodies, divided into two learning batches of 100 variants. 
The mean and variance of each batch is reported.
}
\label{fig:Hist}
\vspace{-.4cm}
\end{figure}

To ensure a fair comparison, we generated 200 sequences for each method and compared them. The results are summarized in Figure \ref{fig:Hist}. 
The antibody variants were partitioned into three equal, chronological batches, starting with the initial dataset, followed by two subsequent batches, each containing 100 mutants. We ensured that all hyperparameters were consistent across methods. The specific hyperparameters can be found in Appendix~\ref{subsec:HyperPar}.

We start with the initial dataset, which has a mean of 69.83 and variance of 14.86. For Bioinformatics-directed evolution, the first batch has a mean of 74.87 (+7.2\%) and the second batch a mean of 75.21 (+7.7\%).
For MLDE with DNN , the first batch has a mean of 87.69 (+25.5\%) and the second batch 93.95 (+34.5\%).
For RCC MLDE with XGBoost, the first batch shows a mean of 86.89 (+24.4\%) and the second batch a mean of 90.47 (+29.5\%). 
For RCC MLDE with DNN, the first batch has a mean of 84.85 (+21.5\%) and the second batch 92.02 (+31.8\%).

Consider sequences in the extreme right tail of the predicted docking-score distribution in Figure~\ref{fig:Hist}.
With a single surrogate trained on pooled conformations, some of these extremes can have high mean scores even though their conformations disagree strongly, so apparent performance depends heavily on which pose is realized.
Empirically, RCC yields populations with high mean docking scores and fewer extreme right-tail points, so top-ranked candidates are strong performers without being dominated by single, idiosyncratic conformations, providing a more conservative and reproducible basis for downstream screening.


RCC introduces a modest additional cost: instead of a single committee, we train one committee per rank, so the cost scales linearly in the number of ranks $R$. In our experiments we use 
$R=4$ ImmuneBuilder conformations per sequence, which is the default samples returned. Computational time remained a small fraction of the overall docking time (which is dominated by HADDOCK3). 

\vspace{-.2cm}
\section{Conclusion and Outlook}\label{sec:discussion}
\vspace{-.2cm}

This paper presents a novel MLDE framework for antibody optimization, addressing the challenges of modeling uncertainty from conformational flexibility and model uncertainty.
By integrating ensemble docking with ImmuneBuilder and applying machine learning models within a Bayesian active learning framework, we show that the method more effectively prioritizes binding candidates in silico compared to traditional single-structure docking.
A current limitation is that final scoring could be further refined with more detailed binding simulations, and additional wet-lab experimental validation will be essential to properly characterize real-world utility.
Nevertheless, our results suggest that this approach offers a scalable route toward the rapid discovery of therapeutic antibodies, particularly in the face of rapidly evolving pathogens such as SARS-CoV-2.

\section*{Acknowledgments}
This work was supported in part by the National Science Foundation under grant DMS-1925919.

\bibliographystyle{plain}
\bibliography{references}

\appendix
\section{Background}\label{app:background}

\paragraph{\emph{In Silico} Scoring.} As traditional DE still incurs high costs from wet-lab evaluation, computational tools based on machine learning have emerged to replace portions of the DE workflow with \textit{in silico} counterparts. Linking such tools together makes it possible to bring the protein design problem \emph{in silico} and use the aforementioned ZO algorithms for protein optimization. A vast amount of software has been developed in recent years to model protein interaction \textit{in silico}, circumventing the traditionally tedious processes of NMR and X-ray crystallography for determining protein complex structure \cite{ritchie2008recent}. This includes HADDOCK3, which incorporates ambiguous interaction restraints to compute different possible docking configurations \cite{honorato2021structural, honorato2024haddock2}, and AutoDock, which specifies flexibility and rigidity in different components of the interaction \cite{goodsell2021autodock}. PRODIGY \cite{xue2016prodigy} computes binding affinity based on structural contact features. We note that different software compute different measures of ``good" binding---for example, while HADDOCK models physical ``fit" and computes an aggregation of energy terms as its final score (estimating $\Delta H$), AutoDock and PRODIGY estimate the binding free energy ($\Delta G$).


\paragraph{Biologically Informed Mutagenesis.} To analyze the relationship between related proteins in terms of function and evolutionary history, biologists have used sequence alignment-based approaches to identify corresponding regions of amino acids. Such approaches rely on amino acid substitution matrices, which reflect the frequencies at which amino acids undergo substitution mutations naturally in proteins over time. Standard substitution matrices include PAM \cite{PAM}, which captures substitution rates among similar sequences, and BLOSUM \cite{henikoff1992amino, henikoff1993performance}, which detects distant relationships by considering substitution rates across dissimilar clusters of sequences. Such biologically derived substitution matrices guide lab-induced mutagenesis with real-world observations of which changes are realistic and stable, as opposed to random sampling with equal likelihood \cite{adenot1999peptides}.


\paragraph{Zeroth-Order (ZO) Learning.} As previously mentioned, it is impossible to test all possible sequences in $S$, due to the combinatorial vastness of the search space. In general, DE can be considered as an \textit{active learning} technique, since new inputs (untested variants) are proposed based on known data (performance of previously tested sequences). Active learning and cluster learning methods have successfully been integrated with wet lab directed evolution to significantly improve properties like product yield and fitness \cite{qiu2021cluster, yang2025active}.

Moreover, zeroth-order optimization strategies like Bayesian optimization have been applied to constrained regions of antibodies to improve binding \cite{khan2022antbo}. Significant progress has been made as well to design appropriate embeddings for CDR regions \cite{singh2025learning} to treat antibody sequences as vectors, which allow sequences to be analyzed more easily in a mathematical framework and incorporated into kernels for active learning techniques. 

Our project builds on these approaches to create a fully end-to-end \emph{in silico} pipeline for antibody optimization through directed evolution using an acquisition function, inspired by zeroth-order optimization. Additionally, with a search space encompassing all three VH CDR regions, our project provides a more comprehensive exploration of possible sequences compared to previous work.

\paragraph{The Sequence-Function Relationship.} The structure of an antibody (and all other proteins) depends on its amino acid sequence, which is encoded in DNA. There are 20 canonical amino acids, each one carrying distinct electrostatic charges, which guide how a protein folds. The resulting folded structure affects how proteins interact with other molecules and ultimately, determine its overall function. It is crucial for protein engineering to understand the sequence-function relationship: finding a protein with an ideal function amounts to finding the ideal amino acid sequence that will produce such function.

\paragraph{Directed Evolution (DE).} When the goal is not to design a protein \textit{de novo}, but rather to improve the function of a known sequence, we can take advantage of helpful starting points for optimizing $h$. Naturally, protein functions are refined throughout thousands or even millions of years of evolution, through random mutation. Scientists can compress this process in the lab with DE. In this iterative process, the scientist induces a mutation in an amino acid sequence and tests the resulting protein's performance on a targeted function. The sequence is either discarded, if the protein's performance is suboptimal, or introduced to a new round of mutation and evaluation \cite{arnold1993engineering}. Simulating Darwininan evolution in the lab allows the researcher to isolate protein function, free of additional biological constraints imposed by the organism from which the protein originated \cite{arnold1996directed}. With scientific software for simulated protein design and evaluation becoming more available, \textit{in silico} directed evolution promises to be a much more cost and time effective process for optimizing our function $h$ that we are trying to minimize . A description of the software used in this project and their capabilities, inputs, and outputs can be found in Data Generation Section.

\paragraph{Zeroth-order (ZO) Optimization.} Directed evolution can conveniently be viewed mathematically as a discrete ZO algorithm. ZO is concerned with optimizing black-box functions without a well-behaved or easily computable gradient \cite{liu2020primer}, as in our case (more details are provided in Chapter \ref{ch:Pipeline}). ZO distinguishes itself from gradient-free and derivative-free optimization methods in that it does not model the gradient $\nabla{h}$ or attempt to extend the discrete-valued function $h$ to a continuous version. Rather, it relies solely on individual evaluations of $h$. 

ZO algorithms in computer science include evolutionary algorithms \cite{wierstra2014natural} and Bayesian optimization \cite{frazier2018tutorial}. Evolutionary algorithms are analogous to the directed evolution procedure in a wet lab. They select new points in the domain to evaluate that are ``mutated" versions of points previously found to have low objective function values. Bayesian optimization utilizes surrogate models of the objective function and an acquisition function to determine high-value points to sample next.

\section{Bioinformatics Section}
\label{sec:Bioinformatics1}

\subsection{Substitution Matrix-based Variant Generation}
{In nature, substitution mutations happen more frequently between some pairs of amino acids than others due to their physio-chemical properties. To faithfully mimic natural mutations, we needed a stochastic matrix $P$ with}
$$P_{ab}=\mathbb P(\text{mutating to aa }b | \text{original aa }a).$$
By extracting the CDR regions from our initial dataset, we established the foundational alignments for our statistical analysis and subsequent construction of the substitution matrix, following the approach of Henikoff et al. \cite{henikoff1993performance}. 
In particular, we calculated substitution frequencies across homologous CDR regions using the Needleman-Wunsch alignment algorithm \cite{murata199022} to create a custom matrix that emphasizes biologically relevant mutations. From our curated CDR alignments, we counted every observed amino acid substitution background frequencies, and normalized the results to create a probabilistic framework for residue changes. The described stochastic matrix reflects the true mutational landscape of antibody hypervariable loops and defines the probability distribution that guides our sequence updates.

The mutation step in our workflow turns random variation into a guided search using two complementary strategies. First, in substitution mutation, we randomly select a CDR region, choose a position within it, and replace a single amino acid based on precomputed probability matrices. Second, with a fixed probability, we apply the crossover mutation, which recombines segments from different sequences. This allows the algorithm to escape local minima and explore entirely new areas of the sequence space.

After the new sequences are processed by our statistical and biophysically enhanced workflow, we advance them through the sequence evaluation process. To mimic evolution, we remove two standard deviations below the mean of sequences (those with the lowest HADDOCK scores) in each iteration, effectively eliminating poorly performing candidates from the sequence pool. 
 

\subsection{Bioinformatics}
\label{sec:Bionformatics2}
The primary aim of Bioinformatics is to narrow the vast sequence space by filtering out the sequences that do not require further analysis. Since all the proceeding algorithm is initiated from the same analysis, we started by exploring and cleaning the sequence dataset received from Avery (Jewett et al. (2023) \cite{hunt2023rapid}). We used Riot-NA \cite{riot} to annotate full-length VH sequences and precisely isolate the three complementarity-determining regions (CDR-H1, H2, H3). By focusing solely on these regions, we addressed the key challenge of reducing the sequence space while retaining the most critical information from the antibody: the areas where binding with the target occurs. Riot-NA matches each sequence to a germline database to find the exact CDR regions. Then, we use strict filters to keep only the clear, high-quality CDRs for further analysis.

After a new mutation is produced, it goes through two steps of verification. Firstly, we verify that the new sequence is biologically feasible and can be produced experimentally. Inspired by \cite{csi-greifflab_developability_profiling} and \cite{sabpred_tap}, we introduced biophysical developability constraints to ensure that each mutated sequence is both synthetically accessible and capable of folding and binding effectively. Our algorithm enforces four principal criteria for developability: hydrophobicity (to promote the burial of hydrophobic residues within the antibody core), aromatic content (to enhance CDR stability and binding through aromatic interactions), charge (to avoid extreme net charges that can lead to aggregation or misfolding), and sequence patterns (to eliminate undesirable motifs such as N‑glycosylation sites or stretches of more than five identical residues). 

The second step in our evaluation is structural verification using AlphaFold2, a state-of-the-art deep learning model that predicts full atomic protein structures from primary sequences \cite{AlphaFold2021}. Each candidate sequence is processed to generate high-confidence 3D models along with per-residue confidence metrics (pLDDT, pTM, ipTM). As the project evolved, AlphaFold2 was replaced with ImmuneBuilder, which, while lacking explicit cutoff metrics, employs an ensemble-based prediction strategy that effectively mitigates structural variability and improves robustness in downstream analysis.

\section{Data Generation}
\label{sec: DataGen}

We build a fully autonomous, end‑to‑end \textit{in silico} pipeline by seamlessly integrating multiple independent, open‑source tools—Riot‑NA, ImmuneBuilder, PBDFixer, and HADDOCK3. Our Scoring Workflow starts with each antibody sequence being first analyzed by Riot‑NA to delineate the heavy and light chain CDR loops. The trimmed sequences are then fed into ImmuneBuilder, which outputs both per‑residue confidence scores and full atomic 3D models. Any gaps or missing atoms in these models are automatically repaired by PBDFixer, ensuring a complete structure for downstream analysis. Finally, the corrected structures are sent to HADDOCK3, where physics‑based docking simulations produce quantitative docking scores that inform the selection of the most promising variants.

We use the SARS‑CoV‑2 antibody database compiled by Jewett et al.\cite{hunt2023rapid}, which catalogs experimentally characterized variable‑region sequences.
Notably, the raw dataset provides improperly annotated CDR regions, which
we re-annotate using Riot-NA \cite{riot}. Accurate identification of CDRs is essential for evaluating how sequence changes may affect protein–protein interactions, which is the foundation of our mutation and scoring workflow.

In earlier versions of the pipeline, each candidate sequence was processed with AlphaFold2 \cite{AlphaFold2021} to generate predicted protein structures and associated confidence metrics (pLDDT, pTM, ipTM). Variants not meeting predefined thresholds for these scores were discarded prior to docking. This step was later replaced by ImmuneBuilder \cite{abanades2023immunebuilder}, which predicts antibody structures with state-of-the-art accuracy using deep learning models trained specifically on immune system proteins.  The output is a set of four different predicted conformations, each in PDB format, which specifies the three-dimensional coordinates of each amino acid residue. Additionally, ImmuneBuilder outputs root mean square deviation (RMSD) values for each amino acid, which, intuitively, quantifies how far a given predicted amino acid is from the average predicted position of that amino acid across the four outputted structures.

We perform the docking step using HADDOCK3 \cite{honorato2024haddock2}, a data-driven platform for modeling biomolecular complexes. HADDOCK3 integrates experimental data and physicochemical parameters to generate structural models of protein-to-protein interactions. For each antibody-antigen pair, the software produces multiple docked complexes in PDB format, along with detailed scoring information that includes van der Waals, electrostatics, desolvation energy, and an overall HADDOCK binding score. The final HADDOCK binding score is a weighted sum of energy terms, designed to approximate the quality and stability of the predicted protein–protein complex. A more negative score indicates stronger binding for a given complex, which is why we employ a minimization function in our set up. HADDOCK3 groups resulting models  into clusters based on structural similarity, and only top-scoring clusters are retained for downstream analysis. This procedure enables us to prioritize variants with strong predicted binding ability and plausible interaction interfaces.

\section{Experimental Procedures}\label{app:exp}

\subsection{Methodology}
\begin{figure}[!ht]
  \centering
\includegraphics[width=1\linewidth]{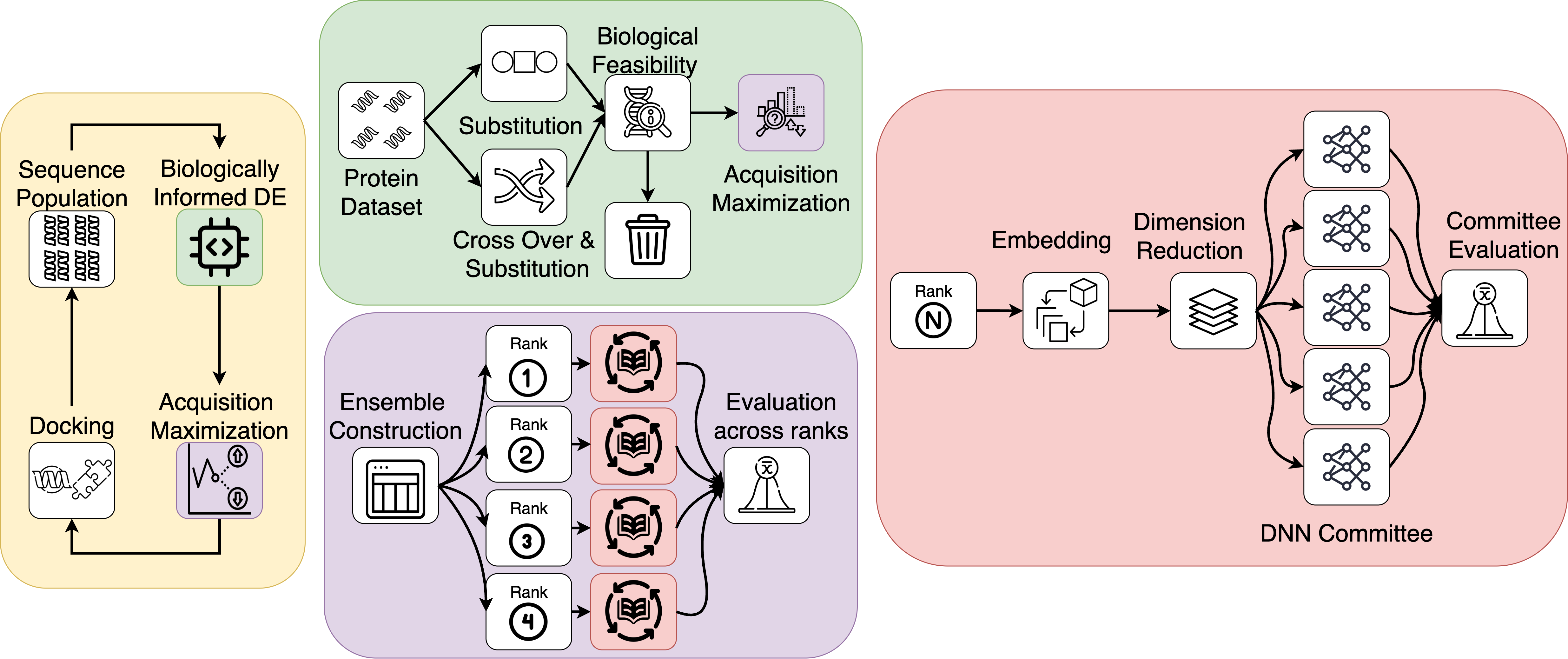}   
\caption{\textcolor{yellow}{\textbf{Yellow Figure}}: General pipeline of the ML-assisted Directed Evolution (MLDE) framework, starting with sequence embeddings and leading to population updates via Acquisition Maximization. \textcolor{green}{\textbf{Green Figure}}: Acquisition Maximization algorithm showing sequence mutations, biological feasibility tests, and sorting by the Acquisition Function to update the population. \textcolor{red}{\textbf{Red Figure}}: Predictive modeling workflow, including PCA-based dimensionality reduction and ensemble modeling using DNN for sequence evaluation. \textcolor{purple}{\textbf{Purple Figure}}: Novel Conformation Rank Committee for acquisition calculation, using ensemble models to predict docking scores and calculate statistical metrics for sequence poses.
}
\label{fig:Pipeline}
\end{figure}

Figure~\ref{fig:Pipeline} illustrates the overall MLDE workflow. Candidate sequences are embedded (yellow panel) and predictive models are trained using ensembles of DNNs and XGBoost (red panel). Sequence libraries are generated by stochastic, biologically informed mutation, filtered for feasibility, and prioritized using an acquisition maximization strategy (green panel). Our contribution is the rank-conditioned committee (purple panel), in which each ImmuneBuilder conformation rank is assigned its own committee, allowing us to aggregate predictions across ranks while separating epistemic and conformational uncertainty. This acquisition strategy enables principled exploration–exploitation balancing without over-penalizing candidates whose uncertainty arises primarily from structural heterogeneity.

\subsection{Hyperparameter Selection in Directed Evolution and Active Learning Approaches}
\label{subsec:HyperPar}

The following hyperparameters were chosen and utilized in the Directed Evolution (DE) and Active Learning (AL) processes during our experiments:

\begin{itemize}
    \item Number of variants generated in each DE loop: 45
    \item Number of loops in Active Learning approaches: 10
    \item Cutoff in each DE loop: 10\% — 90\% of the best-performing sequences are carried forward to the next loop
    \item Selection of top sequences: After each set of 450 generated sequences, 50 sequences with the best acquisition function scores were selected for the next stage
\end{itemize}

For the Acquisition function, we used the Upper Confidence Bound (UCB) approach with a Kappa value of 2, which balances exploration and exploitation of the sequence space.

For the DE Bioinformatics Substitutions, we employed an equal probability of cross and substitution mutations, as outlined in prior literature.

\subsection{More on Ensemble Docking}

\begin{figure}[!ht]
  \centering
    \begin{tabular}{c|c}
    
    \includegraphics[width=.45\textwidth]{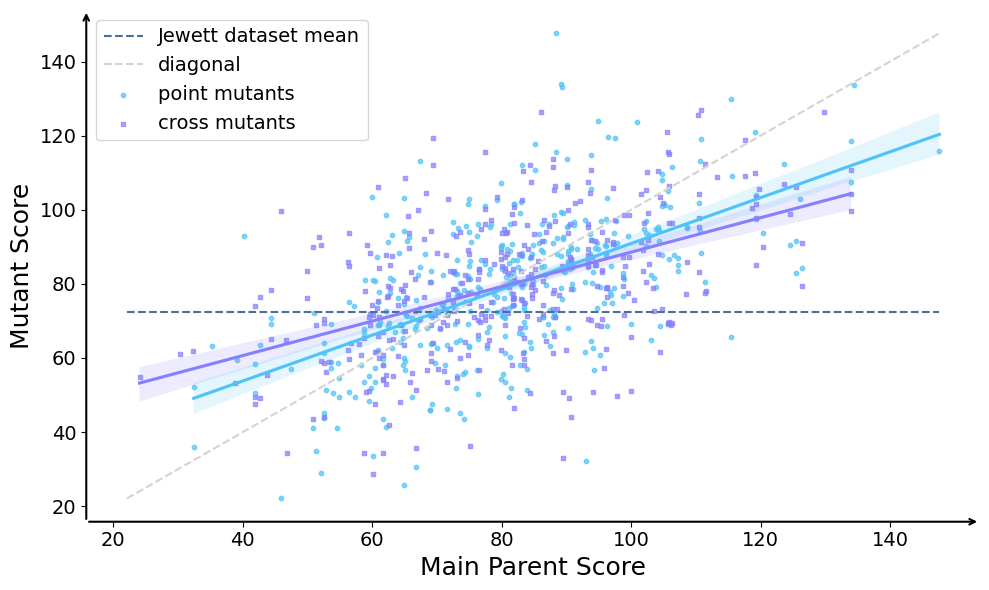}
         &  
    \includegraphics[width=.45\textwidth]{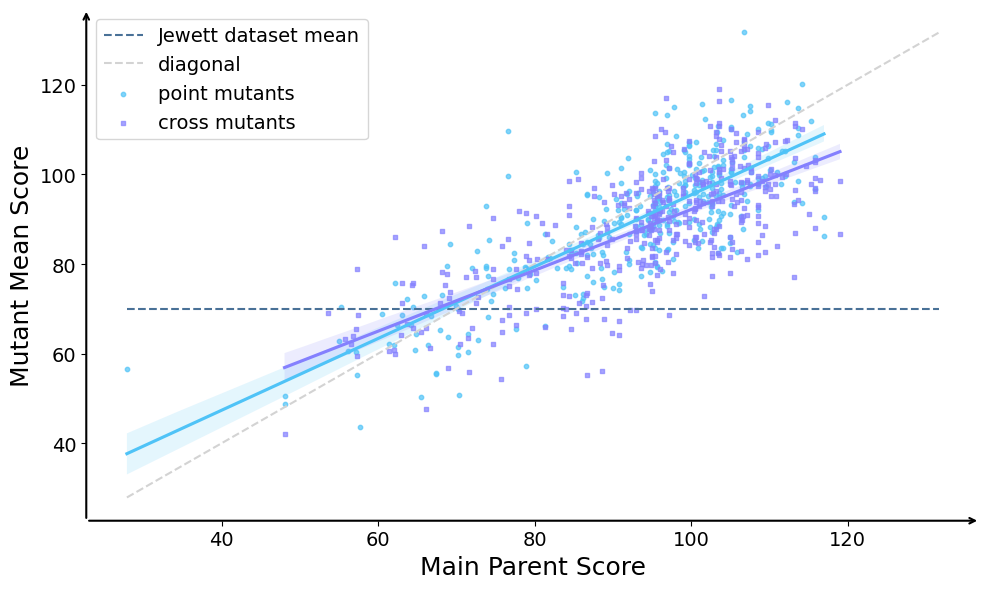}
         \\
        (a): single docking with AlphaFold2 & 
        (b): ensemble docking with ImmuneBuilder
    \end{tabular}
  \caption{Regression plots for mutant scores against main parent scores in both pipelines. 
  Point mutants (light blue) and cross mutants (purple) are shown against main parent scores. Dashed lines: mean of initial population (dark blue) and identity line (grey). Regression results are summarized in Tables \ref{tab:pipeline-a} and \ref{tab:pipeline-b}. In both pipelines, slopes for point mutants exceed those for cross mutants, indicating more moderate impact of point mutations.}
  \label{fig:regression}
\end{figure}

To explore how parent and mutant scores relate, Figure \ref{fig:regression} plots mutant scores against their main parent scores. We fitted separate linear regressions for point and cross mutants (Tables \ref{tab:pipeline-a} and \ref{tab:pipeline-b}), finding a strong positive correlation in both cases in both pipelines ($p<0.001$). In pipeline (b), the slope for point mutants (0.8011) exceeds that for cross mutants (0.6790), suggesting that point mutations have a more moderate impact on scores. The same phenomenon is also discovered in pipeline (a), but its respective slopes and $r^2$ are lower.

\begin{table}[ht]
\centering
\small
\begin{tabular}{@{}lrrrrrr@{}}
\toprule
Term & Coefficient & Std.\ Error & $t$ & $p$ & \multicolumn{2}{c}{95\% CI} \\
\cmidrule(lr){6-7}
     &             &             &     &     & Lower & Upper \\
\midrule
Intercept            & 35.8273 & 4.813 &  7.444 & $<0.001$ & 26.365 & 45.289 \\
Main parent score    &  0.4602 & 0.041 & 11.138 & $<0.001$ &  0.379 &  0.541 \\
Side parent score    &  0.0811 & 0.044 &  1.846 & 0.066    & -0.005 &  0.167 \\
\bottomrule
\end{tabular}
\caption{Pipeline (a): single docking with AlphaFold2 ($r^2=0.251$).}
\label{tab:pipeline-a}
\end{table}

\vspace{1em}

\begin{table}[ht]
\centering
\small
\begin{tabular}{@{}lrrrrrr@{}}
\toprule
Term & Coefficient & Std.\ Error & $t$ & $p$ & \multicolumn{2}{c}{95\% CI} \\
\cmidrule(lr){6-7}
     &             &             &     &     & Lower & Upper \\
\midrule
Intercept            & 19.5007 & 3.154 &  6.183 & $<0.001$ & 13.302 & 25.700 \\
Main parent score    &  0.5391 & 0.044 & 12.277 & $<0.001$ &  0.453 &  0.625 \\
Side parent score    &  0.1906 & 0.042 &  4.533 & $<0.001$ &  0.108 &  0.273 \\
\bottomrule
\end{tabular}
\caption{Pipeline (b): ensemble docking with ImmuneBuilder ($r^2=0.540$).}
\label{tab:pipeline-b}
\end{table}

\vspace{1em}

For cross mutants, we also ran multivariate regressions including both main and side parent scores (Table \ref{tab:pipeline-a}). In both pipelines, the main parent remains the dominant predictor ((a): coefficient = 0.4602 and $p<0.001$, (b): coefficient = 0.5391 and $p<0.001$), while the side parent contributes a less significant effect ((a): coefficient = 0.0811 and $p=0.066$, (b): coefficient = 0.1906 and $p<0.001$).

Moreover, the $r^2$ of the multivariate linear regression of pipeline (b) (0.540) is more than twice that of pipeline (a) (0.251). By comparing all regression results between two pipelines, we conclude that mutant scores are more correlated with the scores of their parents in pipeline (b), which also forms implicit evidence of the accuracy of ensemble docking.

\section{AI Antibody Experiment}\label{app:we-lab-experiments}

\subsection{Linear Expression Template (LET) generation}
The Linear expression templates (LETs) were designed in SnapGene (Dotmatics) as per mentioned in the literature [1]. Variable heavy and light-chain sequences identified by the AI antibody pipeline were synthesized as double-stranded DNA fragments (gBlocks, Integrated DNA Technologies, Coralville, IA, USA). gBlocks were amplified with Q5 Hot Start DNA Polymerase (New England Biolabs, Ipswich, MA, USA) using primers described in the same study. PCR products were purified using gel electrophoresis and used directly for downstream cell-free expression.
\subsection{Cell-free protein synthesis (CFPS)}
Cell-free protein synthesis (CFPS) reactions were carried out using the NEB PURExpress In Vitro Protein Synthesis Kit, according to the manufacturer’s protocol. Briefly, 25 ng of LET DNA was added to a 12.5 µL PURExpress reaction mixture, which consisted of 5 µL Solution A, 3.75 µL Solution B, and nuclease-free water. Reactions were supplemented with the NEB Disulfide Bond Enhancer to facilitate proper folding. All reactions were incubated at 37 °C for 2 hours.
\subsection{AlphaLISA assay for SARS-CoV-2 RBD binding}
Binding of the sdFab constructs to SARS-CoV-2 receptor binding domain (RBD) was assessed using the AlphaLISA assay. CFPS reactions containing expressed sdFabs were mixed with RBD protein and incubated at room temperature for 1 hour prior to AlphaLISA reaction setup. The AlphaLISA reaction mixture contained anti-FLAG donor beads at a final concentration of 0.08 mg/mL, streptavidin acceptor beads at 0.02 mg/mL, and 3 µL (0.12 v/v) of CFPS-RBD reaction product. These components were prepared in Alpha Buffer, which consisted of 50 mM HEPES at pH 7.4, 150 mM NaCl, 1 mg/mL BSA, and 0.00015 v/v Triton X-100. The total reaction volume was 25 µL, and samples were incubated at room temperature in the dark for 1 hour. Following incubation, 25 µL of each AlphaLISA reaction was transferred to a 96-well Corning half-area opaque plate. Plates were read on a PHERAstar plate reader using the AlphaLISA filter set, with an excitation time of 100 ms, an integration time of 300 ms, and a settling time of 20 ms.

\subsection{Experiment Results \& Discussion:}
\paragraph{Assay principle}
AlphaLISA is a wash free, homogeneous, proximity assay using antibody conjugated donor and acceptor beads. If the donor bead is excited by 680 nm wavelength, it generates a singlet oxygen (O$_2$) which has a finite lifetime. If an immunocomplex is formed with the protein, the emitted singlet oxygen transfers energy to the acceptor bead which results in emission at 615 nm. The magnitude of luminescent signal detected by a plate reader is used to quantify binding of protein with antibody.
\paragraph{Data processing}
The raw signals were background subtracted (plate/blank control) and replicates were averaged. Then the average value was normalized to the literature parent antibody (ID150/SC2-31), which was set to 100 arbitrary units for cross-design comparison.
\subsection{Results}

\begin{figure}
    \centering
        \begin{tabular}{cc}
        \includegraphics[width=0.42\textwidth]{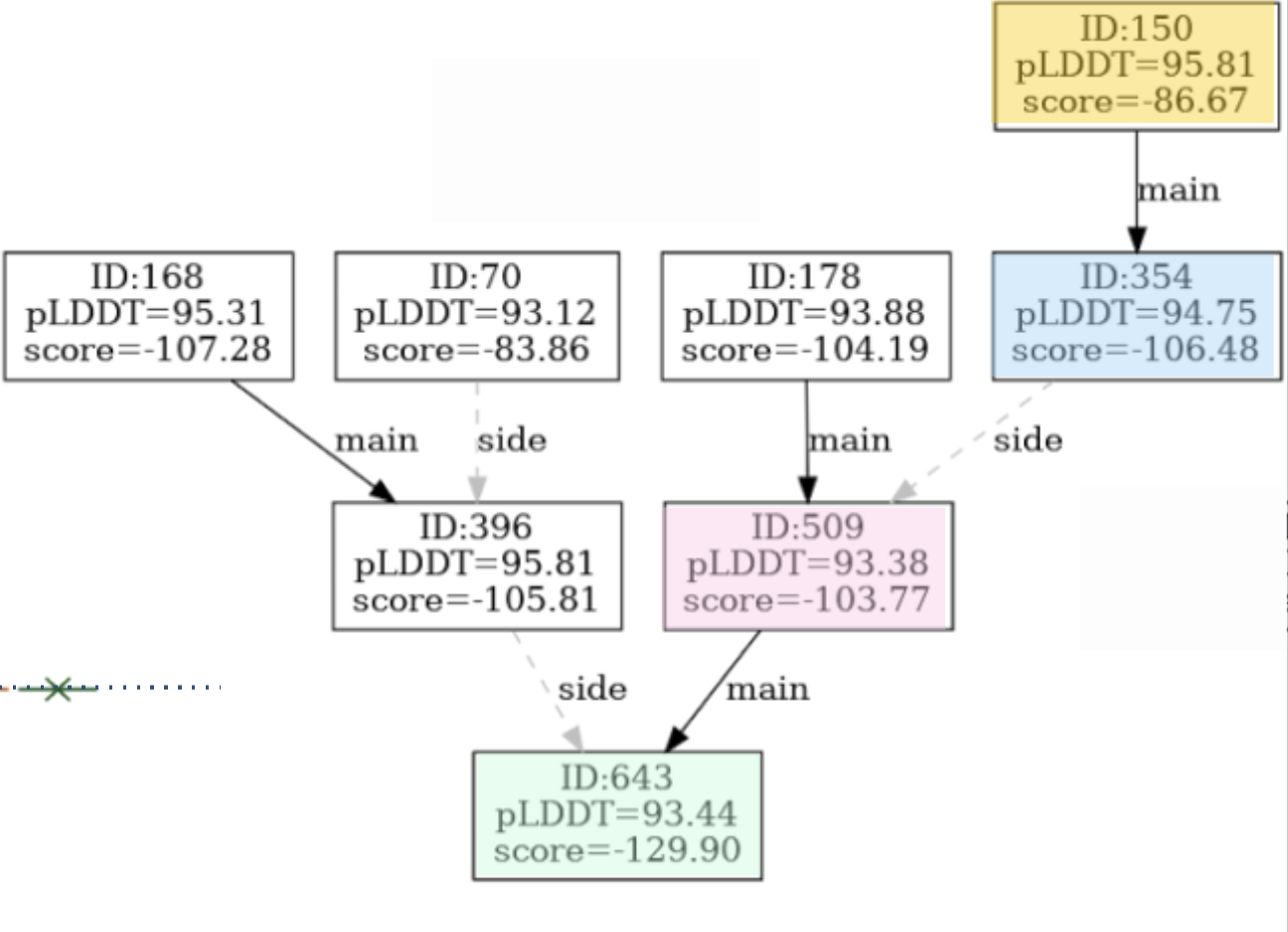}
             &  
        \includegraphics[width=0.5\textwidth]{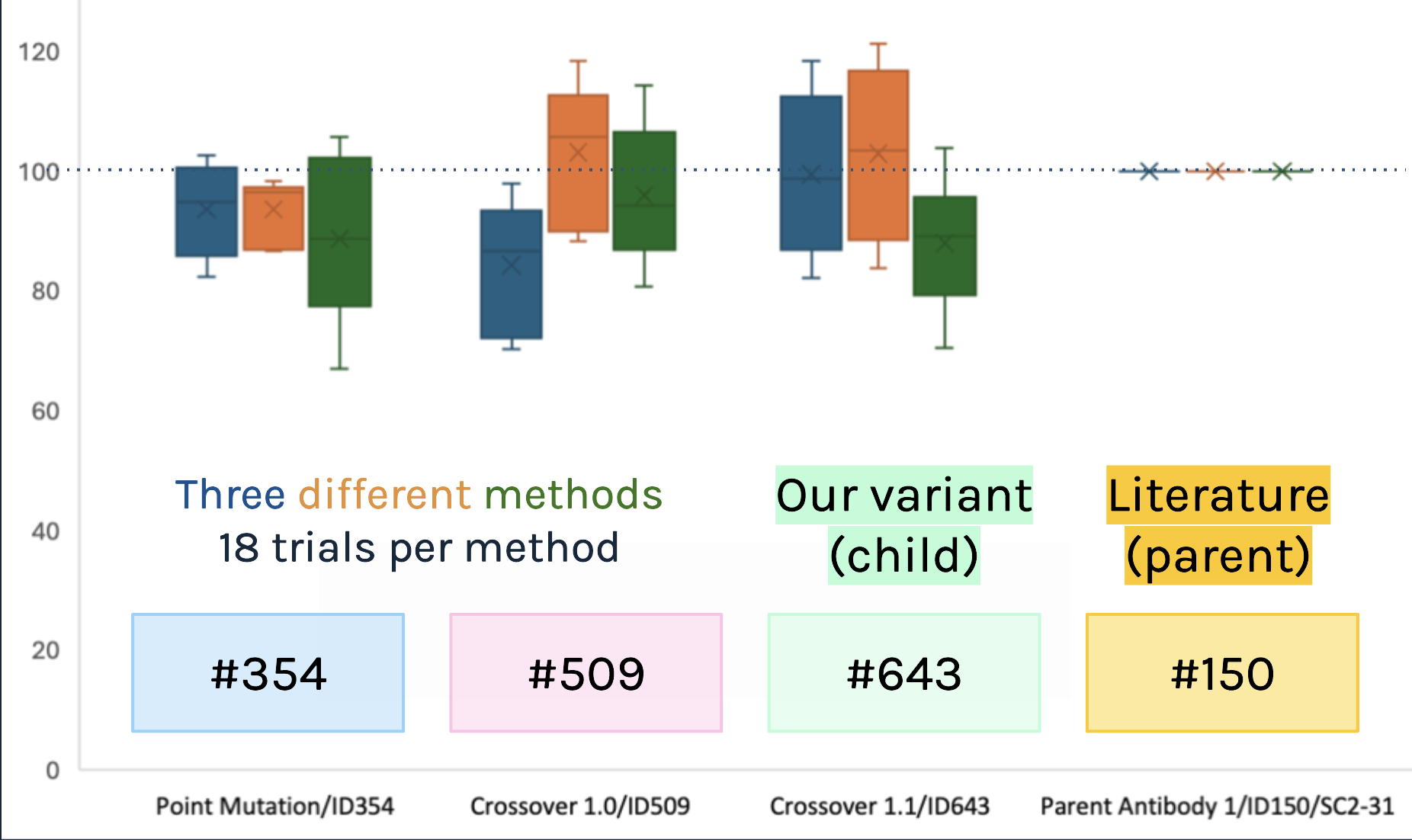}
             \\
             (a) Parent chain & 
             (b) Sidechain Experiment
        \end{tabular}
    \caption{Sidechain genealogy and the resulting parent chain.}
    \label{fig:wet-lab-results}
\end{figure}

\paragraph{Assay performance}
Positive controls yielded robust Alpha signals well above background, confirming effective bead bridging under the stated conditions. Negative controls (buffer only, single-bead controls, or non-binding sdFab) remained near baseline, indicating minimal nonspecific proximity. (Not shown in the graph).
Screening of designed variants
Variants from three computational design routes-point mutation (parent ID354), crossover 1.0 (parent ID509), and crossover 1.1 (parent ID643) were evaluated against the parent (ID150/SC2-31), taken from Jewett et al dataset. Each route was checked with 18 trials, and all signals were normalized to the parent. The crossover 1.1 showed a response distribution centered at or above the parent, w exceeding the parent’s signal. Moreover, the point mutation and crossover 1.0 routes showed broader, mixed distributions result around the parent.
These results are placed illustrated in Figure~\ref{fig:wet-lab-results}.
\subsection{Discussion}
These data demonstrate that AlphaLISA is an effective way to screen the sdFab-RBD binding and our directed evolution framework can yield variants with improved invitro performance relative to a strong literature benchmark. In particular, the crossover 1.1 lineage delivered “child” variants that surpassed the parent in normalized Alpha signal, indicating productive exploration of sequence space.
Also, AlphaLISA reports proximity-based luminescence rather than a direct equilibrium affinity; thus, higher signal generally correlates with stronger binding.

\end{document}